\documentclass{article}

    \PassOptionsToPackage{numbers,sort&compress}{natbib}

\usepackage[main, final]{neurips_2025}

\usepackage[utf8]{inputenc} 
\usepackage[T1]{fontenc}    
\usepackage{hyperref}       
\usepackage{url}            
\usepackage{booktabs}       
\usepackage{amsfonts}       
\usepackage{nicefrac}       
\usepackage{microtype}      
\usepackage{xcolor}         
\definecolor{titlelink}{HTML}{2F5B8C}
\usepackage{fontawesome}
\usepackage{graphicx}       
\usepackage{amsmath}
\usepackage{wrapfig}
\usepackage{multirow}
\usepackage{capt-of}
\newcommand{\rev}[1]{#1}
\newenvironment{revblock}{}{}

\title{\texorpdfstring{\raisebox{-0.04em}{\includegraphics[height=0.82em]{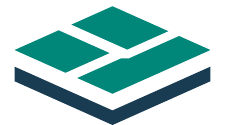}}\hspace{0.25em}}{}MultiTurnRoute: A Platform for Building and Evaluating Multi-Turn LLM Routers}

\author{
Jiaqi Xue\textsuperscript{\texttt{1}}, Mengxin Zheng\textsuperscript{\texttt{1}}, Heng Huang\textsuperscript{\texttt{2}}, Qian Lou\textsuperscript{\texttt{1}}\\
\textsuperscript{\texttt{1}}LoopLeap Lab, University of Central Florida,
\textsuperscript{\texttt{2}}University of Maryland, College Park\\
\texttt{\{jiaqi.xue,mengxin.zheng,qian.lou\}@ucf.edu}\quad
\texttt{heng@umd.edu}\\[4pt]
{\hypersetup{pdfborder={0 0 0}}%
\small\faGithub~Project\enspace\href{https://github.com/loopleap-lab/MultiTurnRoute}{\color{titlelink}\texttt{https://github.com/loopleap-lab/MultiTurnRoute}}}
}

\hypersetup{pdftitle={MultiTurnRoute: A Platform for Building and Evaluating Multi-Turn LLM Routers},pdfauthor={Jiaqi Xue, Mengxin Zheng, Heng Huang, Qian Lou}}
\begin{document}

\maketitle
\vspace{-0.2in}
\begin{center}
\includegraphics[width=\linewidth]{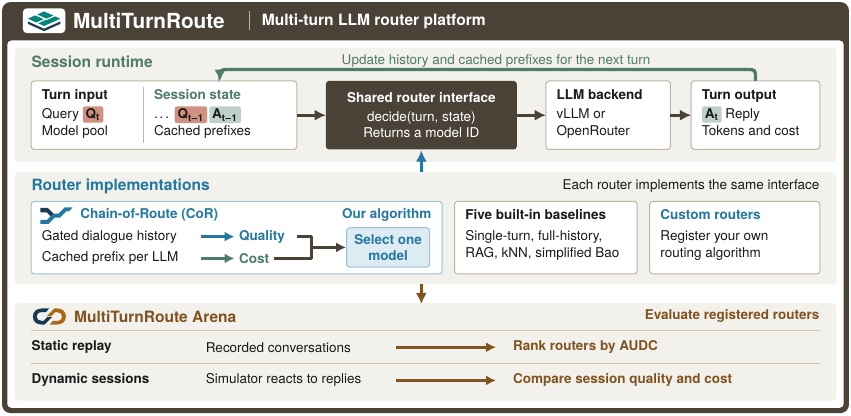}
\captionof{figure}{\textbf{MultiTurnRoute is a multi-turn LLM router platform.} A registered router reads the query and session state to select a model. The reply and token counts update the state for the next turn. The Arena ranks routers by area under the deferral curve (AUDC) in static replay and compares quality and cost in interactive sessions.}
\label{fig:overview}
\end{center}
\vspace{-0.05in}
\begin{abstract}
In a multi-turn conversation, the choice of language model at one turn can affect which model should answer the next. The selected model's reply shapes the context for the next query, and its cached prompt prefix can reduce the cost of using that model again. We introduce MultiTurnRoute, a platform for building and evaluating multi-turn LLM routers. A shared interface gives built-in and user-defined routers the current query, dialogue history, and an estimate of each model's reusable prefix length. Our routing algorithm, Chain-of-Route (CoR), uses a learned gate over recent and recurrent dialogue history to predict response quality and estimates cost from each model's reusable prefix. MultiTurnRoute Arena ranks routers on recorded conversations and compares session quality and cost in interactive runs where a simulated user responds to each reply. On a benchmark drawn from WildChat and LMSYS-Chat-1M with per-turn quality annotations for five LLMs, CoR reduces the normalized cost required to match the strongest model's quality by up to 20.3\% relative to the strongest routing baseline, while matching or improving peak response quality.

\end{abstract}

\section{Introduction}
\label{sec:intro}
A conversational assistant can draw on several large language models (LLMs), each with different capabilities and inference costs. A router decides which model should answer a given query. Most routers make this decision from the current input alone~\cite{ong2024routellm}. In an ongoing conversation~\cite{shi2026lamas}, however, a short follow-up can depend on several earlier turns. In WildChat~\cite{zhao2024wildchat}, a corpus of 1M real user interactions, 41\% of conversations span multiple turns. Other conversation datasets show similar patterns~\cite{sharegpt2023dataset, zheng2023lmsys}. Topics and user goals can also shift within a conversation~\cite{xu2021topic,zhang2020topical}, so the model best suited to one turn may be a poor choice for the next.

We introduce \textbf{MultiTurnRoute}, a platform for building and evaluating routers that select a model at each turn (Figure~\ref{fig:overview}). Its session runtime exposes two kinds of state through a common router interface. The \emph{dialogue state} helps a router interpret the query. The \emph{serving state} tracks the prefix each candidate model is expected to reuse. The selected model's reply extends the dialogue, and its token usage updates the cache record for the next decision.

\begin{figure*}[h!]
    \centering
    \includegraphics[width=\linewidth]{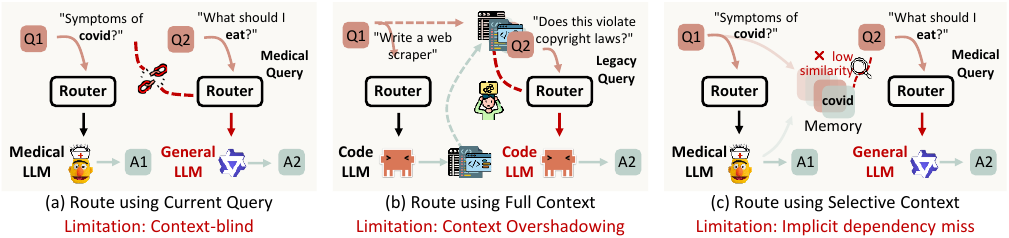}
    \caption{Challenges of adapting single-turn LLM routing to multi-turn conversations regarding dialogue state. (a)~Routing using only the current query ignores conversational context. (b)~Routing using the full context can let dominant prior content overshadow the current query's intent (\textit{context overshadowing}). (c)~Routing using selectively retrieved context misses implicit dependencies when users refer back without repeating keywords.}
    \label{fig:motivation}
    \vspace{-0.08in}
\end{figure*}

\setlength{\columnsep}{8pt}%
\begin{wrapfigure}[10]{r}{0.36\textwidth}
\vspace{-0.17in}
\centering
\includegraphics[width=0.30\textwidth]{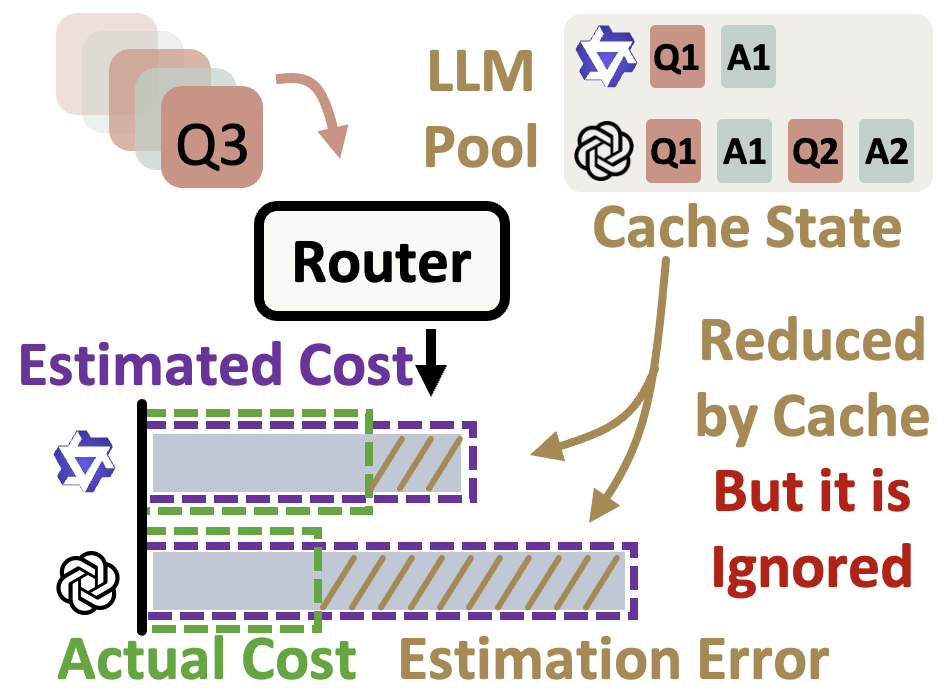}
\vspace{-0.12in}
\caption{Ignoring serving state produces misleading cost rankings.}
\vspace{-0.12in}
\label{fig:motivation_cache}
\end{wrapfigure}

\noindent\textbf{Which history matters for quality?}
After a discussion of COVID symptoms, ``what should I eat?'' asks for dietary advice during illness. A router that sees only the current query can miss this dependency (Figure~\ref{fig:motivation}a). Feeding it the full history introduces \emph{context overshadowing} when, for example, a long code response dominates an unrelated follow-up (Figure~\ref{fig:motivation}b). Retrieving the most similar turns can also miss an implicit reference because the dietary question shares few words with the medical discussion (Figure~\ref{fig:motivation}c). The amount and kind of history needed therefore depend on the current query.

\noindent\textbf{Which prefix matters for cost?}
Serving engines such as vLLM~\cite{kwon2023efficient} and SGLang~\cite{zheng2024sglang} can reuse a model's cached key--value (KV) states for a shared prompt prefix. For example, the cited Claude Sonnet prices are \$3.00 per 1M input tokens and \$0.30 per 1M cached tokens, a 90\% reduction~\cite{anthropic2025pricing}. Each model may have answered a different subset of earlier turns and retained a different prefix. In Figure~\ref{fig:motivation_cache}, switching to a model with a lower nominal price costs more than continuing with the model that already holds the session prefix.

\noindent\textbf{Routing algorithm.}
Our router, \textbf{Chain-of-Route (CoR)}, estimates quality and cost separately for each candidate. Its Dialogue State Propagator (DSP) combines recent turns with a compact recurrent history and learns how much context to use for each query. Its Serving State Tracker (SST) uses each model's cached prefix to estimate the cost of answering. On WildChat and LMSYS-Chat-1M, CoR reduces Query-Normalized Cost (QNC), the minimum normalized cost required to match the strongest model's quality, by up to 20.3\% relative to the strongest routing baseline, while matching or improving peak response quality.

\noindent\textbf{Evaluating routers.}
CoR, five built-in baselines, and user-defined routers implement the same interface. \textbf{MultiTurnRoute Arena} uses that interface in two settings. Static replay compares choices on the same recorded turns and ranks routers by their quality--cost trade-off. In interactive sessions, a simulated user responds to the selected model's reply, so an early routing decision can change the questions a router faces later.

\section{Background and Related Work}

\noindent\textbf{Multi-Turn LLM Conversations.}
Both conversational assistants and multi-agent systems make successive LLM calls over shared context~\cite{xue2026aecp,shi2026lamas,rajib2026gatedmemory}. WildChat~\cite{zhao2024wildchat}, ShareGPT~\cite{sharegpt2023dataset}, and LMSYS-Chat-1M~\cite{zheng2023lmsys} document these interactions. Studies of topic shifts~\cite{xu2021topic,zhang2020topical,budzianowski2018multiwoz} and context dependence~\cite{su2019utterance,wu2017sequential} help explain why different turns may call for models with different capabilities and costs~\cite{ong2024routellm,vayani2025all}. Existing conversation datasets~\cite{zhao2024wildchat, sharegpt2023dataset, zheng2023lmsys,bai2024mt,zheng2023judging} generally record a single model's replies in each dialogue. Routing benchmarks~\cite{hu2024routerbench} provide quality labels for several models on individual queries. Evaluating per-turn routing requires both dialogue histories and ratings for multiple candidate models on each turn.

\noindent\textbf{KV-Cache Management in LLM Serving.}
Successive prompts in a conversation share a prefix, allowing an LLM to reuse key--value states from earlier turns~\cite{shi2026kvsharearena,li2026packcache}. vLLM~\cite{kwon2023efficient} manages these states with paged allocation, SGLang~\cite{zheng2024sglang} supports automatic prefix caching, and CachedAttention~\cite{gao2024cachedattention} extends reuse by offloading states to host memory~\cite{rajib2026agentservesim}. In the cited Claude Sonnet pricing, cached input tokens cost 10\% of the standard input price~\cite{anthropic2025pricing}. Each routing decision affects which model accumulates reusable state. MultiTurnRoute tracks the reusable prefix length for each candidate, and CoR uses it to estimate cost.

\noindent\textbf{LLM Routing.}
Single-turn routers differ in how they predict model quality~\cite{ong2024routellm,hu2024routerbench,feng2024graphrouter,r2router2026} and cost~\cite{song2025irt,shirkavand2025cost,somerstep2025carrot,kabir2026hw,kabir2026inframind} (Appendix~\ref{app:routing_survey}). GMTRouter~\cite{xie2025gmtrouter} learns user preferences across conversations and selects one LLM for a session. Bao et al.~\cite{bao2025dynamic} extend RouteLLM~\cite{ong2024routellm} to choose a model at each turn from the full dialogue history. Their cost model distinguishes switching from staying with the last model. CoR estimates how much of the prompt each candidate retains, including a model used several turns earlier. Concurrent work, MTRouter~\cite{zhang2026mtrouter}, jointly encodes model embeddings and a token-limited interaction history for agentic tool use. Its cost estimate uses token counts and list prices. The authors also observe that switching models frequently can limit prompt-cache reuse.

\section{The Chain-of-Route Algorithm}
\label{sec:method}

\subsection{Overview: From Single-Turn to Multi-Turn Routing}
Let $\mathcal{M} = \{m_1, m_2, \dots, m_K\}$ denote a set of $K$ candidate LLMs. At each turn $t$ of a conversation, the user submits a query $Q_t$ and the router selects a model $m_t$ to generate the answer $A_t$. The router estimates each candidate's response quality $\hat{Q}$ and serving cost $\hat{C}$, then selects the model that maximizes the quality-cost tradeoff, where $\lambda \in [0, 1]$ controls cost sensitivity:
\begin{equation}\small
\label{eq:routing_objective}
    m_t = \arg\max_{m \in \mathcal{M}} \left[(1-\lambda) \cdot \hat{Q} - \lambda \cdot \hat{C} \right].
\end{equation}
Equation~\ref{eq:routing_objective} is the Lagrangian relaxation of maximizing response quality subject to a serving-cost budget, with $\lambda$ acting as the multiplier: sweeping $\lambda$ over $[0,1]$ traces the quality-cost Pareto frontier, and our metrics in Section~\ref{sec:results} summarize this entire frontier rather than any single operating point. In deployment, $\lambda$ is a user-specified quality-cost preference, analogous to the effort levels exposed by commercial assistants. Following recent single-turn routers~\cite{song2025irt,somerstep2025carrot}, we estimate quality and cost separately. Decoupling the two estimates is natural here: KV-cache reuse avoids recomputing an identical prefix, so it changes serving cost but not the response. This lets SST compute the cache correction analytically while DSP focuses entirely on quality.
Single-turn routers estimate $\hat{Q}$ and $\hat{C}$ based solely on the current query $Q_t$: $\hat{Q} = f_Q(m, Q_t)$ and $\hat{C} = f_C(m, Q_t)$. This does not account for two forms of state that accumulate across turns in multi-turn conversations: the \textit{dialogue state}, i.e., the conversation history that shapes query meaning, and the \textit{serving state}, i.e., the KV-cache retained by each model's serving engine that reduces serving cost.

We observe that each routing decision both \textit{depends on} and \textit{shapes} the state available to future decisions. On one hand, the router must consult both forms of state to make an informed choice: the dialogue state determines how to assess each model's quality on the current query, while the serving state determines each model's true serving cost. On the other hand, the routing outcome alters both states for subsequent turns: the selected model's answer extends the dialogue history, and its serving engine retains the KV-cache, reducing its serving cost if chosen again later. This means routing decisions are inherently chained, with state propagating from one turn to the next.

\begin{figure*}[t!]
    \centering
    \includegraphics[width=\linewidth]{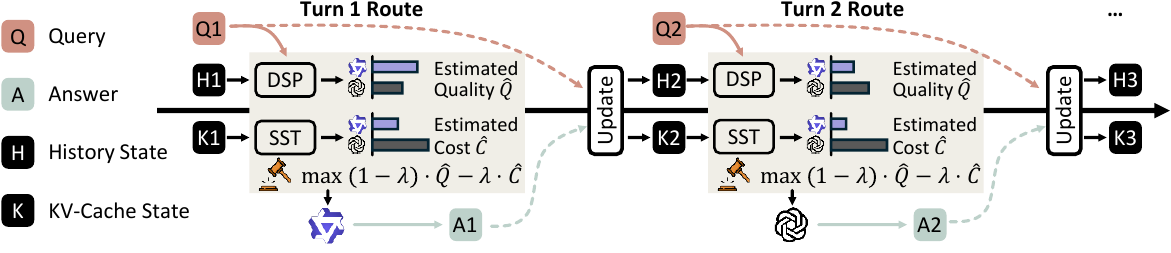}
\caption{Overview of Chain-of-Route (CoR). At each turn, the router receives the current query $Q_t$, the dialogue state $H_t$, and the serving state $\mathcal{K}_t$. DSP conditions on $H_t$ to estimate quality $\hat{Q}$, while SST conditions on $\mathcal{K}_t$ to estimate cost $\hat{C}$. The model maximizing $(1-\lambda)\cdot\hat{Q} - \lambda\cdot\hat{C}$ is selected, and both states are updated before propagating to the next turn.}
\label{fig:technique_overview}
\vspace{-0.2in}
\end{figure*}

Chain-of-Route (CoR), our routing algorithm in MultiTurnRoute, makes this dependence across turns explicit. As illustrated in Figure~\ref{fig:technique_overview}, the router receives the current query $Q_t$ along with two forms of accumulated state: the dialogue state $H_t$ and the serving state $\mathcal{K}_t = \{L_{\text{cache}}^{m,t}\}_{m \in \mathcal{M}}$, where $L_{\text{cache}}^{m,t}$ records the cached token count of candidate model $m$. Two dedicated modules produce state-aware estimates: the Dialogue State Propagator (DSP) conditions on $H_t$ to produce $\hat{Q} = f_Q(m, Q_t, H_t)$, and the Serving State Tracker (SST) conditions on $L_{\text{cache}}^{m,t}$ to produce $\hat{C} = f_C(m, Q_t, L_{\text{cache}}^{m,t})$. These estimates are plugged into Equation~\ref{eq:routing_objective} for model selection.
After the selected model generates its answer $A_t$, both states are updated: DSP incorporates the new query-answer pair to produce $H_{t+1}$, and SST updates each model's cache record to produce $\mathcal{K}_{t+1}$. The updated states carry forward to the next turn, forming a continuous route-then-update chain across the conversation. We detail DSP in Section~\ref{sec:dsp} and SST in Section~\ref{sec:sst}.

\subsection{Dialogue State Propagator (DSP)}
\label{sec:dsp}

As illustrated in Figure~\ref{fig:motivation}, routing based solely on the current query is context-blind~(a), feeding the full history introduces context overshadowing~(b), and selectively retrieving relevant history misses implicit dependencies~(c). Rather than conditioning on raw dialogue at each turn, DSP maintains a compact hidden state $H_t$ that incrementally accumulates conversation-level information (e.g., topic evolution and cross-turn dependencies) through a learned recurrent update~\cite{li2026attend,radi2026keepsake}. This decouples the router from the raw dialogue while preserving the contextual signals necessary for accurate quality estimation.

DSP decomposes $H_t$ into two complementary components: a \textit{local history state} $H_t^{\ell} \in \mathbb{R}^d$ that captures fine-grained, recent context (e.g., the current topic and turn-level difficulty), and a \textit{global history state} $H_t^{g} \in \mathbb{R}^d$ that captures coarse-grained, conversation-level trends (e.g., overall user expertise and long-term topic distribution). The concatenated state $H_t = [H_t^{\ell}; H_t^{g}]$ is used at each turn to inform the routing decision, then updated with the new query-answer pair before propagating to the next turn.

\begin{figure*}[t]
    \centering
    \includegraphics[width=\linewidth]{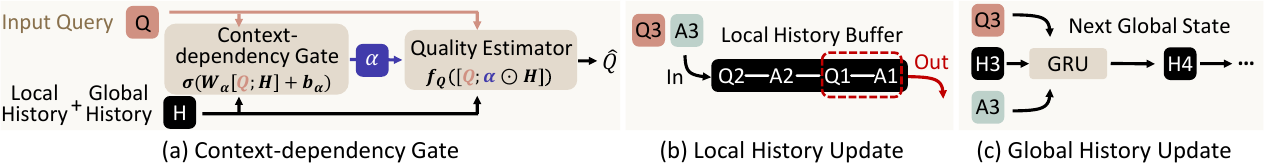}
\caption{Overview of the Dialogue State Propagator (DSP). (a)~Context-dependency Gate: query $Q$ and dialogue state $H$ produce a gate $\alpha_t$, which scales $H$ before the quality estimator $f_Q$. (b)~Local History Update: a FIFO buffer (e.g., of size $W{=}2$) evicts the oldest pair $(Q_1, A_1)$ upon each new arrival $(Q_3, A_3)$. (c)~Global History Update: a GRU updates the global state at each turn (e.g., $H_3 \to H_4$), accumulating conversation-level trends.}
\label{fig:dsp}
\vspace{-0.1in}
\end{figure*}

DSP maintains two states that are read \emph{before} routing each turn and updated \emph{only after} the selected model generates its answer. Concretely, when routing turn $t$, DSP consults the buffer $\mathcal{B}_{t-1}$ and global state $H_t^{g}$ produced after turn $t{-}1$, computes a fresh local state $H_t^{\ell}$ from $Q_t$ and $\mathcal{B}_{t-1}$, forms the dialogue state $H_t = [H_t^{\ell}; H_t^{g}]$, and feeds $H_t$ together with $Q_t$ to a context-dependency gate that controls how much history reaches the quality estimator. After $A_t$ is generated, a frozen pretrained sentence encoder $E$ (all-MiniLM-L6-v2~\cite{reimers2019sentence}, $d{=}384$) produces $d$-dimensional embeddings of the new pair, which we denote $Q_t, A_t \in \mathbb{R}^d$ (overloading the textual symbols), and concatenates them as $x_t = [Q_t; A_t] \in \mathbb{R}^{2d}$ to advance both states to $\mathcal{B}_t$ and $H_{t+1}^{g}$. We use the convention that subscript $t$ on $H_t$ denotes the state available at the start of turn $t$ (i.e., before observing $A_t$). The remainder of this subsection details the three components in the order they execute.

\noindent\textbf{Local History State.}
The local history state maintains a fixed-size FIFO buffer $\mathcal{B}_{t-1} = \{x_{t-W}, \dots, x_{t-1}\}$ of the $W$ most recent past turn embeddings, where $W$ is the local-history window size and each $x_i = [Q_i; A_i]$ corresponds to a completed query-answer pair, i.e., Q and A in Figure~\ref{fig:dsp}~(b). When routing turn $t$, the local state is obtained via cross-attention, with the current query $Q_t$ as the query and the buffered past turn embeddings as keys and values:
\begin{equation}\small
    H_t^{\ell} = \mathrm{CrossAttn}(Q_t,\ \mathcal{B}_{t-1}) = \mathrm{softmax}\!\left(\frac{(Q_t W_Q^{\ell})(\mathcal{B}_{t-1} W_K^{\ell})^\top}{\sqrt{d}}\right)\mathcal{B}_{t-1} W_V^{\ell},
\end{equation}
where $W_Q^{\ell} \in \mathbb{R}^{d \times d}$ and $W_K^{\ell}, W_V^{\ell} \in \mathbb{R}^{2d \times d}$ are learned projections. After the selected model produces $A_t$, the buffer is advanced: $\mathcal{B}_t \leftarrow (\mathcal{B}_{t-1} \cup \{x_t\}) \setminus \{x_{t-W}\}$, with positional order preserved via position embeddings. This allows the router to selectively attend to the most relevant recent turns rather than treating all turns equally.

\noindent\textbf{Global History State.}
The global history state employs a gated recurrent update to selectively accumulate conversation-level information across all turns, as illustrated in Figure~\ref{fig:dsp}~(c). When routing turn $t$, the router reads the current global state $H_t^{g}$ (produced after turn $t-1$) directly. After the selected model generates $A_t$, two gates jointly determine how $H_t^{g}$ is advanced to $H_{t+1}^{g}$ using the new turn input $x_t = [Q_t; A_t]$. The \textit{relevance gate} $r_t$ assesses which dimensions of $H_t^{g}$ are pertinent to the new turn, selectively filtering outdated information:
\begin{equation}\small
    r_t = \sigma(W_r [H_t^{g}; x_t] + b_r).
\end{equation}
The \textit{retention gate} $z_t$ controls the balance between preserving the accumulated state and incorporating new information, learning to make small updates for routine turns while allowing larger updates when the turn carries significant conversation-level signals (e.g., a shift in user expertise level):
\begin{equation}\small
    z_t = \sigma(W_z [H_t^{g}; x_t] + b_z).
\end{equation}
The candidate state $\tilde{H}_t$ is then computed by combining the relevance-filtered history with the new turn input:
\begin{equation}\small
    \tilde{H}_t = \tanh(W_h [r_t \odot H_t^{g}; x_t] + b_h),
\end{equation}
where $\sigma$ denotes the sigmoid function and $\odot$ denotes element-wise multiplication. Finally, the global state is advanced as a weighted combination of the previous state and the candidate:
\begin{equation}\small
    H_{t+1}^{g} = (1 - z_t) \odot H_t^{g} + z_t \odot \tilde{H}_t.
\end{equation}
The global state is initialized as $H_1^{g} = \mathbf{0}$, and the local buffer $\mathcal{B}_0$ is initially empty, so at the first turn the router falls back to query-only routing.

\noindent\textbf{Context-dependency Gate.}
Not all queries in a multi-turn conversation depend equally on prior context. Some are follow-ups that inherit meaning from earlier turns, while others introduce entirely new topics and can be routed independently. This suggests that the router should adaptively decide how much history to incorporate for each query, rather than applying a fixed strategy. With $H_t^{\ell}$ and $H_t^{g}$ in hand, DSP forms the dialogue state $H_t = [H_t^{\ell}; H_t^{g}]$ and computes a \textit{context-dependency gate} $\alpha_t \in [0, 1]$ that assesses whether the current query requires conversational context, as shown in Figure~\ref{fig:dsp}~(a):
\begin{equation}\small
    \alpha_t = \sigma(W_\alpha [Q_t; H_t] + b_\alpha).
\end{equation}
When $\alpha_t \approx 1$, the query is context-dependent and the router leverages the full dialogue state; when $\alpha_t \approx 0$, the query is independent and the router effectively reduces to single-turn routing based on $Q_t$ alone. The quality estimator then incorporates context in proportion to this gate:
\begin{equation}\small
    \hat{Q}(m, Q_t, H_t) = f_Q([Q_t; \alpha_t \odot H_t]),
\end{equation}
where $[\cdot; \cdot]$ denotes concatenation and $f_Q$ is a learned scoring function.

\subsection{Serving State Tracker (SST)}
\label{sec:sst}

\begin{figure*}[t!]
\vspace{0in}
    \centering
    \includegraphics[width=0.98\linewidth]{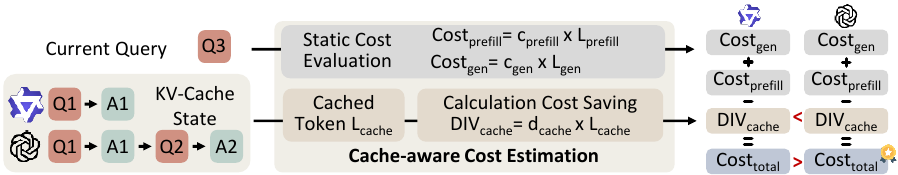}
\caption{Overview of SST. Taking turn~3 as an example, Qwen caches $(Q_1, A_1)$ while GPT caches $(Q_1, A_1, Q_2, A_2)$. (Left)~Each model's serving state and cached token count $L_{\text{cache}}$. (Middle)~Static cost estimation yields a lower estimate for Qwen; cache dividend estimation computes each model's $\text{DIV}_{\text{cache}} = d_{\text{cache}} \times L_{\text{cache}}$. (Right)~After subtracting $\text{DIV}_{\text{cache}}$, GPT's larger cache advantage reverses the ranking, making it the truly more cost-efficient option.}
\label{fig:sst}
\vspace{-0.1in}
\end{figure*}

As discussed earlier, existing routers estimate cost using a static formula: $C(m, Q_t) = c_{\text{prefill}}^m \cdot L_{\text{prefill}}^t + c_{\text{gen}}^m \cdot L_{\text{gen}}^t$, where $c_{\text{prefill}}^m$ and $c_{\text{gen}}^m$ are model $m$'s per-token prices and $L_{\text{prefill}}^t$, $L_{\text{gen}}^t$ are the prefilling and generation token counts. This does not account for KV-cache reuse: a model that has served prior turns benefits from cache hits on previously seen tokens, but the static formula prices every token at the full rate. Since each candidate may have served different prior turns, the true serving cost varies across models. SST addresses this by introducing a \textit{cache dividend} correction to the static cost estimate. In short, SST prices each candidate by how many context tokens it already holds in its KV-cache: cached tokens are charged at the cached-input rate and only the rest at the full prefill rate. We illustrate SST with a concrete example in Figure~\ref{fig:sst}. Qwen answered turn~1. GPT received that exchange as part of its prompt for turn~2 and generated $A_2$. When $Q_3$ arrives at $t{=}3$, Qwen can therefore reuse $(Q_1,A_1)$ and GPT can reuse the longer prefix $(Q_1,A_1,Q_2,A_2)$, assuming both caches persist.

\noindent\textbf{Serving State Tracking.}
SST maintains the cache state $L_{\text{cache}}^{m,t}$ for each candidate model $m$, defined as the number of tokens for which model $m$ currently retains KV-cache at turn $t$. An instrumented serving engine can expose the retained prefix directly. With provider APIs, MultiTurnRoute estimates it from the prompt and output token counts on the most recent turn served by $m$, capped at the current prompt length. Eviction between calls can shorten the actual prefix. The backend's reported cache hits determine the realized input cost when available (Appendix~\ref{app:eviction}). In the running example of Figure~\ref{fig:sst}~(left), Qwen holds $L_{\text{cache}}^{\text{Qwen},3}=|Q_1,A_1|$ while GPT holds $L_{\text{cache}}^{\text{GPT},3}=|Q_1,A_1,Q_2,A_2|$, reflecting their different serving histories.

\noindent\textbf{Static Cost Estimation.}
As shown in Figure~\ref{fig:sst} (middle, top), a state-unaware router prices each candidate by the same static formula based on the full context length $L_{\text{prefill}}^t$: a prefilling cost $\text{Cost}_{\text{prefill}} = c_{\text{prefill}}^m \cdot L_{\text{prefill}}^t$ and a generation cost $\text{Cost}_{\text{gen}} = c_{\text{gen}}^m \cdot L_{\text{gen}}^t$, following the static cost formulation adopted by single-query routing work~\cite{somerstep2025carrot}. Because per-token prices $c_{\text{prefill}}^m$ and $c_{\text{gen}}^m$ differ across candidates, the static estimate already produces a per-model ranking. In our running example, Qwen has lower per-token prices than GPT, so its static estimate is the smaller of the two. Crucially, this ranking is computed as if every candidate must prefill the entire conversation from scratch, ignoring how much of the context each candidate has already cached.

\noindent\textbf{Cache Dividend Estimation.}
SST corrects the static estimate by computing the \textit{cache dividend} $\text{DIV}_{\text{cache}}$, the savings from reusing cached KV states, shown in Figure~\ref{fig:sst}~(middle, bottom). With per-token cache discount $d_{\text{cache}}^m = c_{\text{prefill}}^m - c_{\text{cached}}^m$ (where $c_{\text{cached}}^m$ is the cached-input price set by the provider for black-box APIs, or $\approx 0$ for self-hosted models), the dividend for candidate $m$ is $\text{DIV}_{\text{cache}}(m) = d_{\text{cache}}^m \cdot L_{\text{cache}}^{m,t}$. Subtracting it from the static estimate yields the cache-aware cost:
\begin{equation}\small
\label{eq:sst_cost}
    \hat{C}(m, Q_t, L_{\text{cache}}^{m,t}) = \overbrace{c_{\text{prefill}}^m \cdot L_{\text{prefill}}^t}^{\text{Cost}_{\text{prefill}}} - \overbrace{d_{\text{cache}}^m \cdot L_{\text{cache}}^{m,t}}^{\text{DIV}_{\text{cache}}} + \overbrace{c_{\text{gen}}^m \cdot L_{\text{gen}}^t}^{\text{Cost}_{\text{gen}}}.
\end{equation}
As illustrated in Figure~\ref{fig:sst} (right), although Qwen has a lower static estimate than GPT, GPT enjoys a much larger $\text{DIV}_{\text{cache}}$ because it retains a longer prefix. Subtracting the dividend reverses the cost ranking in this example. A state-unaware router would prefer Qwen based on the static estimate and miss this opportunity.

By integrating DSP and SST, the final routing decision at turn $t$ reduces to Equation~\ref{eq:routing_objective}, where $\hat{Q}$ is the dialogue-state-conditioned quality from DSP and $\hat{C}$ is the serving-state-aware cost from SST, jointly optimizing quality and cost under realistic multi-turn serving conditions.
\section{Evaluation of Chain-of-Route}
\label{sec:results}

\subsection{Experimental Setup}
\noindent\textbf{Model pool.}
We evaluate CoR and the comparison methods on a shared pool of $K{=}5$ models. The Arena protocols are described separately in Section~\ref{sec:multiturnserve}. The experimental pool comprises Qwen3-0.6B, Qwen3-8B, Qwen3-14B, Qwen3-32B, and Qwen3-Next-80B-A3B-Instruct. We take each model's full-prefill price $c_{\text{prefill}}^m$ and generation price $c_{\text{gen}}^m$ from publicly listed OpenRouter pricing, and set the cached-input price as $c_{\text{cached}}^m = 0.2\, c_{\text{prefill}}^m$ following Alibaba Cloud Model Studio's context-cache convention.\footnote{\url{https://www.alibabacloud.com/help/en/model-studio/context-cache}} The full per-model pricing table is in Appendix~\ref{app:cost_model}.

\noindent\textbf{Dataset.}
Existing multi-turn dialogue benchmarks~\cite{zheng2023judging,bai2024mt,zheng2023lmsys} are single-model and do not provide per-turn multi-model quality scores; we therefore construct evaluation data from WildChat~\cite{zhao2024wildchat} and LMSYS-Chat-1M~\cite{zheng2023lmsys}. From each corpus we randomly sample 20{,}000 English conversations ($\sim$80{,}000 turns) with $\geq 2$ turns per conversation, and split 80/20 by conversation hash.

To train the quality predictor, we need per-(turn, candidate) quality labels. We obtain them by having every candidate model $m \in \mathcal{M}$ generate a response for every training turn $t$ under the \emph{same} dialogue history, using the dataset's original responses for prior turns:
\begin{equation}\small
    \text{context}_{\mathrm{train}}(t) = \{(Q_1, A_1^{\text{orig}}), \ldots, (Q_{t-1}, A_{t-1}^{\text{orig}}), Q_t\}.
\end{equation}
Each candidate response is independently scored on a 1--10 scale by DeepSeek-V3.1, selected from a panel of candidate judges via meta-evaluation against human ratings~\cite{zheng2023judging} (details in Appendix~\ref{app:dataset}). \rev{We adopt this shared-context construction as a tractable offline protocol}: faithfully enumerating all routing trajectories scales as $|\mathcal{M}|^T$ in the conversation length, which is intractable. These quality scores supervise DSP. A separate head predicts output length from observed response lengths. SST combines this prediction with the tracked prefix length and token prices to estimate cost (Appendix~\ref{app:implementation}).

At evaluation, each routing strategy follows one trajectory per held-out conversation. For each user query in that conversation, CoR uses DSP and SST to select a model under Equation~\ref{eq:routing_objective}. Each comparison method uses its own decision rule. The selected model generates a reply, which DeepSeek-V3.1 judges and which becomes part of the history for the next query.

\noindent\textbf{Baseline Methods.} \textit{Random} selects uniformly per turn. Single-turn routers \textit{RouteLLM}~\cite{ong2024routellm}, \textit{CARROT}~\cite{somerstep2025carrot}, and \textit{CSCR}~\cite{shirkavand2025cost} route each turn from the current query alone. Multi-turn routers \textit{Bao et al.}~\cite{bao2025dynamic} (full-history input + binary switch penalty) and \textit{GMTRouter}~\cite{xie2025gmtrouter} (cross-conversation user graph) are the closest competitors. \rev{The \textit{Myopic Oracle} in Figure~\ref{fig:results_main}~(a) is a per-turn hindsight oracle: along its own trajectory, it picks the candidate maximizing Equation~\ref{eq:routing_objective} using judge-assigned quality and realized cost under the current states, without branching over future decisions ($O(KT)$ per $\lambda$, not $O(K^T)$).} All learned routers share \textit{all-MiniLM-L6-v2}~\cite{reimers2019sentence} as the text encoder; training hyperparameters, $\lambda$ grid, and baseline adaptations are in Appendix~\ref{app:implementation}.





\noindent\textbf{Evaluation Metrics.} We evaluate each routing strategy using a \textit{deferral curve}~\cite{jitkrittum2025universal,shirkavand2025cost}, which plots the average per-turn response quality against the average normalized cost. Sweeping the cost-sensitivity parameter $\lambda$ over the interval $[0, 1]$ (Equation~\ref{eq:routing_objective}) traces the deferral curve. Following~\cite{jitkrittum2025universal}, we employ three evaluation metrics: (i) Area Under the Deferral Curve (AUDC), measuring the overall quality-cost trade-off; (ii) Peak Quality, the maximum quality achievable by the router; and (iii) Query-Normalized Cost (QNC), the minimum normalized cost required to match the quality of the most capable LLM in the pool. Unless noted otherwise, ablations are run on WildChat with the default configuration; a sensitivity analysis over CoR's main design choices (local-history window, global state architecture, text encoder, cache discount) is provided in Appendix~\ref{app:sensitivity}.

\subsection{Results}

\noindent\textbf{Overall Quality-Cost Tradeoff.}
Table~\ref{tab:main_results} shows that CoR delivers the best quality-cost trade-off on both datasets. Single-turn routers miss both the dialogue state and the serving state. GMTRouter targets cross-conversation personalization (\emph{who} asks rather than \emph{what context} accumulates) and lands near the single-turn cluster. Bao et al., the strongest baseline, feeds full history into a RouteLLM-style router and reduces serving cost to a binary switching penalty; CoR pulls ahead by maintaining a compact dialogue state and tracking per-model cache. The QNC gap is the largest, as SST exploits cache reuse that baselines do not plan for.

\vspace{-0.1in}
\begin{table}[b!]
\centering
\footnotesize
\caption{Quality-cost tradeoff on WildChat and LMSYS-Chat-1M.}
\setlength{\tabcolsep}{8pt}
\renewcommand{\arraystretch}{1.0}
\begin{tabular}{lcccccc}
\toprule
\multirow{2}{*}{Method} & \multicolumn{3}{c}{WildChat} & \multicolumn{3}{c}{LMSYS-Chat-1M} \\
\cmidrule(lr){2-4} \cmidrule(lr){5-7}
 & AUDC $\uparrow$ & QNC $\downarrow$ & Peak $\uparrow$ & AUDC $\uparrow$ & QNC $\downarrow$ & Peak $\uparrow$ \\
\midrule
Random               & $0.49$ & $0.97$ & $0.87$ & $0.55$ & $0.93$ & $0.78$ \\
\midrule
RouteLLM~\cite{ong2024routellm}    & $0.56$ & $0.98$ & $0.94$ & $0.62$ & $0.93$ & $0.83$ \\
CARROT~\cite{somerstep2025carrot}  & $0.59$ & $0.95$ & $0.94$ & $0.65$ & $0.91$ & $0.83$ \\
CSCR~\cite{shirkavand2025cost}     & $0.68$ & $0.85$ & $0.94$ & $0.65$ & $0.87$ & $0.83$ \\
\midrule
GMTRouter~\cite{xie2025gmtrouter}  & $0.65$ & $0.87$ & $0.94$ & $0.62$ & $0.89$ & $0.84$ \\
Bao et al.~\cite{bao2025dynamic}   & $0.70$ & $0.74$ & $0.95$ & $0.67$ & $0.78$ & $0.85$ \\
\midrule
\textbf{CoR (Ours)}                & $0.74$ & $0.59$ & $0.95$ & $0.70$ & $0.65$ & $0.87$ \\
\bottomrule
\end{tabular}
\label{tab:main_results}
\end{table}

\begin{figure}[t]
    \centering
    \includegraphics[width=\linewidth]{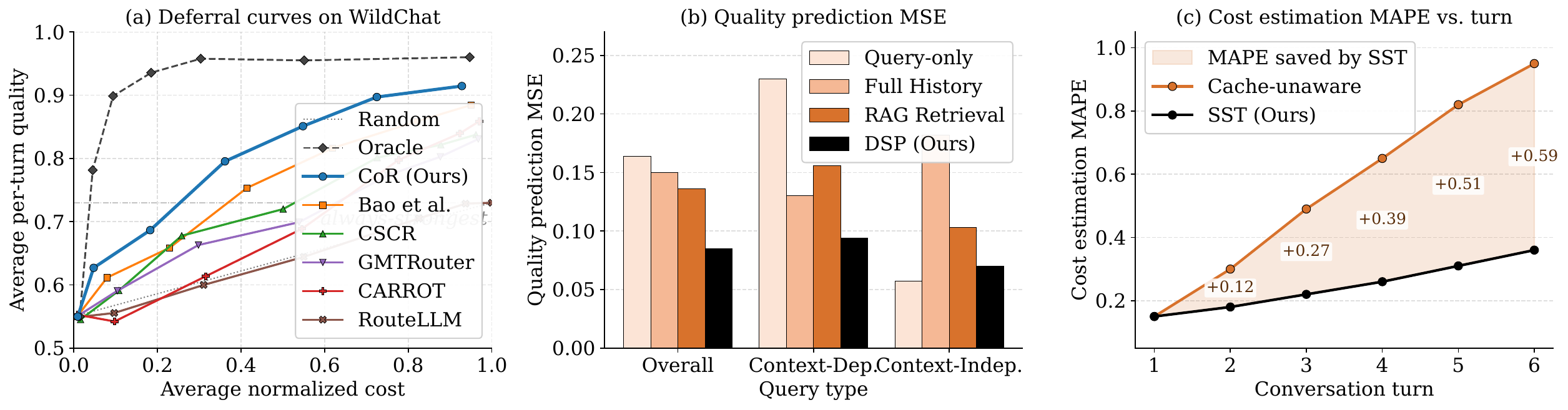}
    \caption{(a) Deferral curves on WildChat as the cost-sensitivity parameter $\lambda$ is swept\rev{; ``Oracle'' is the per-turn Myopic Oracle defined in Section~\ref{sec:results}}. (b) Quality prediction MSE for four history strategies, broken down by query type. (c) Cost estimation MAPE as a function of conversation turn for cache-unaware estimation vs.\ SST.}
    \label{fig:results_main}
\end{figure}

Figure~\ref{fig:results_main}~(a) plots the full deferral curve on WildChat as $\lambda$ is swept over $[0, 1]$. CoR Pareto-dominates all baselines across most of the cost range, with the gap widest at moderate cost budgets where DSP's context-aware selection and SST's cache-aware costing combine to extract the largest savings.

\noindent\textbf{Effect of Dialogue State on Quality Prediction.}
To verify that DSP's recurrent dialogue state improves quality prediction, we compare it against three history-consumption strategies (i.e., Query-only, Full History, and RAG Retrieval) on two kinds of queries: \emph{context-dependent} (where removing all prior turns drops the gold quality score by at least 3 points on the 1--10 scale, a $\geq 30\%$ relative drop) and \emph{context-independent} (the rest). \rev{Under this partition, 58.7\% of held-out WildChat turns and 54.2\% of LMSYS-Chat-1M turns are context-dependent, so dialogue state materially affects quality on the majority of turns.} As shown in Figure~\ref{fig:results_main}~(b), DSP outperforms all baselines on context-dependent queries. Query-only is the weakest on this slice, matching the context-blind failure in Figure~\ref{fig:motivation}~(a), while Full History and RAG Retrieval also underperform, consistent with context overshadowing~(b) and implicit dependencies~(c). On context-independent queries, DSP and Query-only perform comparably, so DSP does not inject harmful signals when history is unnecessary.

\begin{wraptable}{r}{0.29\textwidth}
\vspace{-0.5em}
\centering
\scriptsize
\caption{Context-dependency gate $\alpha_t$ behavior.}
\setlength{\tabcolsep}{3pt}
\renewcommand{\arraystretch}{1.0}
\begin{tabular}{l c}
\toprule
Query type           & mean $\alpha_t$              \\
\midrule
Context-dependent    & $0.78$     \\
Context-independent  & $0.28$     \\
\bottomrule
\end{tabular}
\label{tab:gate_behavior}
\vspace{-0.5em}
\end{wraptable}

Although never supervised with this partition, the gate fires strongly on context-dependent queries (mean $\alpha_t \approx 0.78$) and stays low on context-independent ones ($\approx 0.28$; Table~\ref{tab:gate_behavior}), so DSP learns to modulate history use rather than uniformly mixing it in.

\noindent\textbf{Effect of Serving State on Cost Estimation.}
Figure~\ref{fig:results_main}~(c) shows cost estimation mean absolute percentage error (MAPE) versus conversation turn. At turn~1, both methods start with the same MAPE ($\sim 0.15$); this floor reflects unavoidable output-length prediction uncertainty (both methods predict generation tokens before observing them), and any cost estimator inherits it. As turns increase, cache-unaware estimation grows steeply and approaches $\sim 0.95$ by turn~6 because it prices every token at the full prefill rate and ignores cached reuse. SST grows only mildly with turn depth (the residual error reflects imperfect cache-state tracking under realistic serving variability) and stays within a narrow band well below cache-unaware. The widening gap shows that cache-unaware estimation increasingly overestimates cost in longer conversations, biasing routing toward unnecessary switches, while SST keeps the bias bounded and enables more efficient routing.
\begin{table}[h!]
\centering
\footnotesize
\caption{Quality-cost performance stratified by conversation length on WildChat. Buckets group test conversations by total turn count. CoR's gain over the strongest single-turn baseline (CSCR) widens as conversations grow longer, consistent with SST exploiting more accumulated cache.}
\setlength{\tabcolsep}{8pt}
\renewcommand{\arraystretch}{1.0}
\begin{tabular}{l c cc cc cc}
\toprule
\multirow{2}{*}{Turn bucket} & \multirow{2}{*}{\#conv.} & \multicolumn{2}{c}{AUDC $\uparrow$} & \multicolumn{2}{c}{QNC $\downarrow$} & \multicolumn{2}{c}{Peak $\uparrow$} \\
\cmidrule(lr){3-4} \cmidrule(lr){5-6} \cmidrule(lr){7-8}
                & & CSCR & CoR & CSCR & CoR & CSCR & CoR \\
\midrule
2 turns         & $\sim$1{,}680 & $0.69$ & $0.71$ & $0.86$ & $0.65$ & $0.96$ & $0.94$ \\
3--4 turns      & $\sim$1{,}240 & $0.67$ & $0.74$ & $0.85$ & $0.58$ & $0.94$ & $0.95$ \\
5--7 turns      & $\sim$\phantom{0}700   & $0.65$ & $0.75$ & $0.88$ & $0.53$ & $0.92$ & $0.96$ \\
8+ turns        & $\sim$\phantom{0}380   & $0.63$ & $0.78$ & $0.92$ & $0.45$ & $0.90$ & $0.97$ \\
\midrule
$\Delta$ (8+ vs 2) & --- & $-0.06$ & $+0.07$ & $+0.06$ & $-0.20$ & $-0.06$ & $+0.03$ \\
\bottomrule
\end{tabular}
\label{tab:turn_stratification}
\end{table}
\vspace{-0.1in}

The downstream effect appears in Table~\ref{tab:turn_stratification}. As conversation length increases, CoR's AUDC rises from $0.71$ (2 turns) to $0.78$ (8+ turns), while CSCR's falls from $0.69$ to $0.63$. The QNC gap widens from $0.21$ to $0.47$. This divergence indicates that SST benefits from cache accumulation: longer conversations provide more reuse, whereas cache-unaware methods repeatedly incur full prefill cost.

\noindent\textbf{Cross-dataset transfer.}
Training on WildChat and evaluating on LMSYS yields AUDC $0.68$, and the reverse direction (LMSYS $\to$ WildChat) gives $0.71$, only $0.02$--$0.03$ below in-domain results. This suggests DSP learns transferable conversation-level signals rather than dataset-specific features.

\noindent\textbf{Candidate-pool composition.}
Varying pool size $K \in \{2,3,4,5,6\}$ and replacing Qwen3 members with Llama-3.1 and Mistral models in a cross-family ablation, CoR remains the best method in all settings, and cross-family performance matches the same-family pool, indicating no reliance on family-specific shortcuts. See Appendix~\ref{app:pool_scalability}.

\rev{\noindent\textbf{Additional analyses.} CoR's gains hold under cross-family judges, dialogue-length and topic shift, and real commercial billing (Appendices~\ref{app:judge}, \ref{app:ood}, and~\ref{app:billing}).}

\vspace{-0.1in}
\section{The MultiTurnRoute Platform}
\label{sec:multiturnserve}

\subsection{Session Runtime and Router Interface}
MultiTurnRoute connects a conversational application to a pool of LLMs (Figure~\ref{fig:overview}). At each turn a registered router reads the current query, dialogue history, and the estimated reusable prefix length for each candidate. It selects one model, and the runtime dispatches the request to its backend. The backend (vLLM~\cite{kwon2023efficient} or a provider API) returns the reply and reports billed and cached tokens when available. The runtime records the exchange and updates the session state. Applications can use an OpenAI-compatible endpoint, including deployments behind gateways such as vLLM Semantic Router~\cite{semanticrouter2025}.

CoR and five built-in baselines (single-turn, full-history, RAG, kNN, and simplified Bao et al.) implement \texttt{decide(turn, state)}. A researcher can register another router with that interface and use the same implementation in the runtime and Arena. The router can use either part of the state. For example, a cache-aware policy can compare the cost of returning to a previously used model with that of continuing on the current model. A history-based policy can use the same input to estimate which model will answer a follow-up well. The runtime executes either choice and accounts for the realized cost using the backend's input, cached, and output token counts. This lets researchers change the decision rule while using the same backend and accounting code. The implementation is available at \url{https://github.com/loopleap-lab/MultiTurnRoute}.

\vspace{-0.05in}
\subsection{MultiTurnRoute Arena}
Single-turn arenas such as RouterArena~\cite{lu2025routerarena} can look up a pre-scored answer for each query. In a conversation, a model's reply can also change what the user asks next. MultiTurnRoute Arena measures these effects in two tracks.

\noindent\textbf{Static replay.}
The static track uses 3{,}996 conversations drawn from the benchmark in Section~\ref{sec:results}. A separate pool of nine models answered every recorded turn in its original context, and a judge rated each answer. On the held-out split, the Arena looks up the rating of the model selected by the router. All routers see the same user queries and recorded replies. Their costs can still diverge because each router's sequence of model choices determines which prefixes its models can reuse. The Arena holds the model pool, prices, and cost weights fixed and ranks routers by area under the deferral curve (AUDC). The leaderboard therefore compares policies within this pool. The main experiments in Section~\ref{sec:results} use their own model pool and quality annotations.

\noindent\textbf{Interactive sessions.}
The dynamic track starts routers from the same task. A simulated user reads the selected model's reply and writes the next question or ends the conversation. An answer that requests clarification, for example, can lead to a different sequence of later turns than an answer that completes the task immediately. A judge rates each answer in its own dialogue context, and the Arena computes cost from the token usage reported by the backend. Routers run on matched tasks with the same model pool and user simulator. For each cost weight, the Arena averages answer quality and normalized cost over rated turns and sums the billed cost across all turns.

\section{Conclusion}
MultiTurnRoute gives researchers a common interface for building multi-turn LLM routers and comparing their decisions on recorded and interactive conversations. Our routing algorithm, Chain-of-Route, uses DSP to estimate response quality from dialogue history and SST to estimate cost from each model's cached prefix. On WildChat and LMSYS-Chat-1M, CoR reduces QNC by up to 20.3\% relative to the strongest baseline while matching or improving peak response quality. Ablations show that both modules contribute. MultiTurnRoute Arena lets researchers compare other routing policies under the same conditions and examine how a model choice affects later turns.

\section*{Acknowledgements}
This work was supported in part by the National Science
Foundation (NSF) under Award No. 2523407. H.H. was partially supported by NSF IIS 2347592, 2348169, DBI 2405416, CCF 2348306, CNS 2347617, RISE 2536663. 
Any opinions, findings, and conclusions or recommendations expressed in this material are those of the author(s) and do not
necessarily reflect the views of the National Science Foundation.

\bibliographystyle{unsrt}
\bibliography{example_paper}

\newpage
\appendix
\setcounter{topnumber}{2}  

\section{Survey of Single-Turn LLM Routers}
\label{app:routing_survey}

LLM routing has been widely adopted in open-source platforms~\cite{huggingface_llm_router,semanticrouter2025} and commercial deployments such as OpenAI's GPT-5~\cite{openai2025gpt5} and Azure AI Foundry~\cite{azure-model-router-2025}. Existing routers vary primarily in how they predict per-candidate quality and cost, but uniformly treat each query in isolation.

\noindent\textbf{Quality predictors.}
Lightweight models are trained to estimate each candidate's response quality via neural networks~\cite{ong2024routellm,chen2024routerdc,vsakota2024fly}, $k$-nearest neighbors~\cite{hu2024routerbench,stripelis2024tensoropera,wu2025efficient}, and graph neural networks~\cite{feng2024graphrouter}.

\noindent\textbf{Cost predictors.}
Early routers rely on static attributes such as per-token pricing~\cite{song2025irt,shirkavand2025cost}, while more recent work incorporates output-length prediction~\cite{nguyen2024metallm,somerstep2025carrot,wu2025efficient} for more accurate per-query cost estimates.

\noindent\textbf{Complementary routing signals.}
A separate line of work introduces complementary signals into the routing decision while still routing each query independently: Arch-Router~\cite{tran2025archrouter} aligns routing with human preferences over latency-quality trade-offs, R2-Router~\cite{r2router2026} introduces reasoning into the routing decision, \rev{HW-Router~\cite{kabir2026hw} and INFRAMIND~\cite{kabir2026inframind} incorporate real-time hardware and serving-system signals such as queue lengths, KV-cache usage, and recent latency to meet latency SLOs,} and past-performance retrieval~\cite{generalising_routing2026} retrieves past performance records to generalize routing to unseen models.

\rev{\noindent\textbf{Safety-aware routing.}
Beyond quality and cost, foundation models~\cite{li2025gvd,radi2026aeroworld} raise safety and security concerns. They are vulnerable to trojan and backdoor attacks that hijack their outputs through hidden triggers~\cite{xue2023trojllm,zheng2024trojfsp,zheng2024sslcleanse} or bias their predictions against specific groups~\cite{xue2024badfair,xue2024trojfair}, to poisoned retrieval corpora in retrieval-augmented generation~\cite{xue2026badragidentifyingvulnerabilitiesretrieval}, and to adversarial text perturbations~\cite{lou2024crutp}. Tracing the provenance of model-generated text calls for robust watermarking~\cite{xue2026pro,alghanim2025evaluating}, and protecting user data during training and inference motivates cryptographic privacy-preserving computation~\cite{zhang2025cipherprune,xue2026robpi,xue2025dictpfl,santriaji2025dataseal,xue2024cryptotrain,xue2026sok,yudha2024boostcom,kumar2026fhecoder} and trusted execution environments~\cite{xue2025securing}. Routing offers a natural point at which to enforce such safeguards~\cite{zhang2026securerouter}: SafeRoute~\cite{lee2025saferoute} routes each input between a small and a large safety guard model, invoking the large model only on hard inputs to keep guardrail accuracy high at lower cost. Like the routers above, it decides for each input independently.}

None of the above account for dialogue or serving state across turns; \rev{CoR explicitly and jointly models both. We compare against Bao et al.~\cite{bao2025dynamic}, which routes each turn of a conversation, as a baseline}.

\section{Dataset Construction Details}
\label{app:dataset}

\noindent\textbf{Why existing multi-turn benchmarks are inadequate.}
MT-Bench~\cite{zheng2023judging} contains only 80 two-turn dialogues, each completed by a single LLM. MT-Bench-101~\cite{bai2024mt} scales to 1{,}388 multi-turn dialogues but remains single-model. LMSYS Chatbot Arena~\cite{zheng2023lmsys} provides large-scale human preferences at the conversation level, not per-turn quality scores. None of these provides what per-turn routing requires: for each turn $t$, quality scores $Q(m, Q_t)$ from \textit{multiple} candidate models under the \textit{same} dialogue context.

\noindent\textbf{Preprocessing.}
For both WildChat and LMSYS-Chat-1M we apply identical preprocessing: filter to English conversations with $\geq 2$ turns and $\geq 10$ characters per user query, then randomly sample 20{,}000 conversations ($\sim$80{,}000 turns) per corpus.

\noindent\textbf{Shared-context protocol for training-label construction.}
The shared-context design used to build the per-(turn, candidate) training labels ensures: (1)~fair comparison---all candidate models see identical context at each turn, so quality differences reflect model capability alone; (2)~\rev{tractability---it yields $K$ comparable labels per turn in $O(KT)$ generation and judge calls, without depending on a particular routing policy}; and (3)~extensibility---adding a new model requires only generating and scoring its responses without regenerating existing data. This shared-context construction is restricted to label generation: at evaluation time we run auto-regressive routing, where each routing strategy traces a single trajectory per held-out conversation and the routed model's actual response is fed back as context for the next turn, faithfully reflecting deployment conditions including any response-distribution shift induced by routing-driven model switches.

\noindent\textbf{LLM judge selection.}
To select a reliable LLM judge, we followed the LLM-as-a-judge validation protocol~\cite{zheng2023judging}: we randomly sampled $500$ candidate responses, and recruited a pool of $30$ expert annotators with each response receiving independent ratings from $3$ annotators (yielding $1{,}500$ human ratings whose mean is the gold label). We then evaluated $4$ candidate judges (DeepSeek-V3.1, Qwen3-235B-A22B, GPT-5-mini, and Llama-3.3-70B-Instruct) by computing their Pearson correlation with the gold labels. Among them, DeepSeek-V3.1 achieved the highest correlation ($\rho=0.84$) and was selected as the final judge. This LLM-as-a-judge approach is crucial for evaluating open-ended dialogue queries, where traditional automatic evaluation methods such as exact match fail to capture response quality.

\noindent\textbf{Scoring details.}
Each model's response at each turn is independently scored on a 1--10 scale. We use individual (non-pairwise) scoring for scalability ($K$ evaluations per turn vs.\ $\binom{K}{2}$ for pairwise) and compatibility with our quality predictor $\hat{Q}$, which estimates absolute quality rather than relative preference. Dialogue context is truncated to 1{,}000 characters per prior message to manage judge token cost; this caps prompt length while preserving topical and stylistic cues that drive multi-turn quality. Scores are returned in structured JSON format with reasoning.

\section{Per-Model Pricing Table}
\label{app:cost_model}

Each candidate model $m$ has three distinct per-token unit prices in our cost model: full-prefill $c_{\text{prefill}}^m$ (paid for tokens that miss the cache), cached-input $c_{\text{cached}}^m$ (paid for tokens served from KV-cache), and generation $c_{\text{gen}}^m$ (paid per output token). The serving-state-aware cost in Section~\ref{sec:sst} multiplies each by the corresponding token count $L_{\text{prefill}}^t$, $L_{\text{cache}}^{m,t}$, $L_{\text{gen}}^t$. We take $c_{\text{prefill}}^m$ and $c_{\text{gen}}^m$ directly from publicly listed OpenRouter pricing for each model. The cached-input price is not separately listed for these open-source Qwen3 models, so we set $c_{\text{cached}}^m = 0.2\, c_{\text{prefill}}^m$ (i.e., an $80\%$ discount), following the cache-discount convention documented by Alibaba Cloud Model Studio.\footnote{\url{https://www.alibabacloud.com/help/en/model-studio/context-cache}} Table~\ref{tab:api_pricing} reports the resulting per-token prices.

\begin{table}[h]
\centering
\scriptsize
\caption{Per-token prices for the five Qwen3 models in our pool, in USD per 1M tokens. Full-prefill ($c_{\text{prefill}}$) and generation ($c_{\text{gen}}$) prices are taken from publicly listed OpenRouter pricing (snapshot retrieved during paper preparation); the cached-input price is set to $c_{\text{cached}} = 0.2\, c_{\text{prefill}}$ following Alibaba Cloud Model Studio's context-cache convention. Note that posted prices are non-monotone in model size (e.g., Qwen3-32B at \$0.16/M is cheaper than Qwen3-14B at \$0.35/M) because providers set prices by tier rather than by size; we use posted prices as-is rather than imputing a monotone schedule.}
\setlength{\tabcolsep}{6pt}
\renewcommand{\arraystretch}{1.05}
\begin{tabular}{l c c c c}
\toprule
Model & Total params & $c_{\text{prefill}}$ & $c_{\text{cached}}$ & $c_{\text{gen}}$ \\
\midrule
Qwen3-0.6B                     & 0.6B  & \$0.11 & \$0.022 & \$0.42 \\
Qwen3-8B                       & 8B    & \$0.18 & \$0.036 & \$0.70 \\
Qwen3-14B                      & 14B   & \$0.35 & \$0.070 & \$1.40 \\
Qwen3-32B                      & 32B   & \$0.16 & \$0.032 & \$0.64 \\
Qwen3-Next-80B-A3B-Instruct    & 80B   & \$0.15 & \$0.030 & \$1.20 \\
\bottomrule
\end{tabular}
\label{tab:api_pricing}
\end{table}

\rev{\noindent\textbf{Cost accounting in evaluation.} Cost accounting is identical for all methods. Equation~\ref{eq:sst_cost} is used only by CoR at routing time. For evaluation, every method's cost is recomputed from its \emph{realized} cache-hit tokens, uncached tokens, and output tokens with the unit prices in Table~\ref{tab:api_pricing}, so baselines receive full credit for any cache reuse.}

\section{Implementation Details}
\label{app:implementation}

\noindent\textbf{Encoder and training.}
All learned routers (CoR and baselines) use \textit{all-MiniLM-L6-v2}~\cite{reimers2019sentence} as the shared text encoder ($d{=}384$). Routing models are trained on the 80\% training split with AdamW (lr $1{\times}10^{-4}$, batch size 64, 20 epochs, early stopping on validation MSE). The training objective is mean-squared error between predicted quality $\hat{Q}(m, Q_t, H_t)$ and the judge-assigned ground-truth quality, summed across all $K{=}5$ candidates.

\noindent\textbf{DSP configuration.}
The local-history window is $W{=}2$, and the global state dimension is $d{=}384$ (matching the encoder).

\noindent\textbf{Cost normalization and $\lambda$ sweep.}
We divide each turn's raw cost (token-weighted by per-model unit cost) by the maximum-model cost so that costs lie in $[0, 1]$. Output length per candidate is predicted by a lightweight regression head following CARROT~\cite{somerstep2025carrot}: a per-candidate linear head on the same MiniLM-L6 query embedding, trained on the 80\% training split with mean-squared error against the candidate's actual response length and reused at routing time. We sweep $\lambda \in \{0, 0.05, 0.10, \ldots, 1.0\}$ when computing the deferral curve.

\noindent\textbf{Baseline adaptations.}
\textit{Bao et al.}~\cite{bao2025dynamic} originally route between two models with a binary switching penalty; we adapt their formulation to our five-model pool by treating each pairwise switch under the same penalty schedule. \textit{GMTRouter}~\cite{xie2025gmtrouter} originally learns per-user preferences from cross-conversation histories; we instantiate it by treating each conversation hash as a distinct user, using the authors' reference implementation. \textit{History-strategy variants} (Query-only, Full History, RAG Retrieval) share CoR's encoder and training pipeline but differ in how they construct the input to the quality predictor. \rev{\textit{Full-history + SST baselines} feed the concatenated dialogue history to CARROT, CSCR, and IRT-Router, keep each router's own quality predictor, and replace its cost estimator with CoR's SST. The \textit{BERT full-history} variant fine-tunes a BERT-base encoder (109.5M parameters) end-to-end on the full dialogue history.}

\noindent\textbf{Component ablation training.}
To ablate DSP or SST individually, we re-train CoR with the corresponding module replaced by its single-turn or cache-unaware counterpart while keeping the other module unchanged.

\section{Compute Resources}
\label{app:compute}

\noindent\textbf{Routing module training.}
CoR's routing modules consist of a frozen \textit{all-MiniLM-L6-v2} encoder ($\sim$22M parameters, kept fixed) plus lightweight per-candidate task heads. Each training configuration trains for at most 20 epochs over the 80\% training split with batch size 64 and completes in under 2 hours on a single NVIDIA A100 (40GB) GPU. The full sensitivity sweep (Appendix~\ref{app:sensitivity}: window $W$, global-state architecture, encoder, cache discount; Appendix~\ref{app:pool_scalability}: pool size $K$; Appendix~\ref{app:eviction}: eviction rate $p_{\mathrm{evict}}$) totals roughly $30$ training runs and approximately 60 GPU-hours.

\noindent\textbf{Candidate-response generation.}
Generating per-(turn, candidate) responses across the five Qwen3 models on $\sim$160K turns (WildChat + LMSYS-Chat-1M) is performed via the OpenRouter API. The resulting shared-context responses are generated once and can be reused for training. Held-out evaluation generates the replies selected along each routing trajectory.

\noindent\textbf{LLM-judge scoring.}
DeepSeek-V3.1 scores each shared-context candidate response on a 1--10 scale. This one-time labeling cost scales with $|\text{turns}| \times K$. The same judge scores replies generated during held-out evaluation. The judge meta-evaluation (500 samples $\times$ 4 candidate judges, Appendix~\ref{app:dataset}) is a small fraction of the total judge spend.

\noindent\textbf{Evaluation.}
Auto-regressive evaluation runs each routing strategy as a single end-to-end trajectory per held-out conversation. Per-strategy evaluation across the held-out test split with $|\lambda|{=}21$ values completes in under 1 hour on the same A100 GPU; total wall-clock is dominated by candidate inference latency rather than the routing decision itself, which is sub-millisecond per turn.

\section{Sensitivity to Design Choices}
\label{app:sensitivity}

This appendix provides the full sensitivity analysis referenced in the main text. Table~\ref{tab:sensitivity} sweeps four key design axes of CoR on WildChat. \textbf{(a)} The local-history window $W$ has a clear optimum at $W{=}2$: $W{=}1$ underfits short-range coreference, while $W{\geq}3$ admits diluting attention from less relevant turns. AUDC varies by at most $0.02$ across the sweep, indicating mild sensitivity. \textbf{(b)} The global state architecture matters: the gated GRU update outperforms a plain RNN ($\Delta$AUDC $+0.03$) and matches a single-layer Transformer at a fraction of the compute, supporting the choice of a recurrent rather than purely attentional aggregator. \textbf{(c)} CoR is largely insensitive to the text encoder: substituting BERT-base or E5-small for MiniLM-L6 changes AUDC by $\leq 0.01$, with E5-small giving a small additional gain. \textbf{(d)} The cache-discount sweep is the most consequential. As $d_{\text{cache}}^m / c_{\text{prefill}}^m$ shrinks from $0.95$ (near-zero cache cost, vLLM-style) to $0.5$ (small dividend), AUDC drops by $0.04$ and QNC rises from $0.55$ to $0.78$. CoR remains the strongest method across all settings, but absolute gains are correlated with how much the deployment platform actually rewards cache reuse, an empirically observable property of the serving stack.

\begin{table}[t]
\centering
\footnotesize
\caption{Sensitivity of CoR to design choices on WildChat. Default configuration in \emph{italics}.}
\setlength{\tabcolsep}{6pt}
\renewcommand{\arraystretch}{1.0}
\begin{tabular}{l cc}
\toprule
Variant & AUDC $\uparrow$ & QNC $\downarrow$ \\
\midrule
\multicolumn{3}{l}{\textit{(a) Local-history window size $W$ (DSP)}} \\
$W = 1$                   & $0.72$ & $0.65$ \\
\emph{$W = 2$ (default)}  & $0.74$ & $0.59$ \\
$W = 3$                   & $0.74$ & $0.61$ \\
$W = 5$                   & $0.73$ & $0.64$ \\
\midrule
\multicolumn{3}{l}{\textit{(b) Global state architecture (DSP)}} \\
\emph{GRU (default)}      & $0.74$ & $0.59$ \\
RNN                       & $0.71$ & $0.68$ \\
Linear Attention          & $0.73$ & $0.62$ \\
Transformer (1-layer)     & $0.74$ & $0.60$ \\
\midrule
\multicolumn{3}{l}{\textit{(c) Text encoder}} \\
\emph{all-MiniLM-L6 (default)}  & $0.74$ & $0.59$ \\
BERT-base                       & $0.74$ & $0.61$ \\
E5-small-v2                     & $0.75$ & $0.58$ \\
\midrule
\multicolumn{3}{l}{\textit{(d) Cache discount $d_{\text{cache}}^m / c_{\text{prefill}}^m$ (SST)}} \\
$0.5$ (small dividend)          & $0.71$ & $0.78$ \\
\emph{$0.8$ (default)}          & $0.74$ & $0.59$ \\
$0.95$ (vLLM near-zero)         & $0.75$ & $0.55$ \\
\bottomrule
\end{tabular}
\label{tab:sensitivity}
\end{table}

\section{Candidate-Pool Composition}
\label{app:pool_scalability}

This appendix expands the candidate-pool composition study summarized in the main text. Table~\ref{tab:pool_scalability} reports the full quality-cost metrics for pool sizes $K \in \{2, 3, 4, 5, 6\}$ over Qwen3 members, and a cross-family pool at $K{=}5$ that replaces three Qwen3 members with Llama-3.1 and Mistral models. CoR's AUDC ranges from $0.78$ at $K{=}2$ to $0.73$ at $K{=}6$, declining gracefully as harder routing decisions are introduced while QNC stays below $0.62$ across all pool sizes. The cross-family pool at $K{=}5$ matches the same-family $K{=}6$ result on AUDC ($0.73$) within rounding, indicating that DSP's quality estimator and SST's cache tracker do not exploit Qwen-family-specific shortcuts.

\begin{table}[t]
\centering
\scriptsize
\caption{Robustness to candidate-pool composition on WildChat. \textit{Top:} pool size $K$ varied from 2 to 6 over Qwen3 models. \textit{Bottom:} cross-family pool at $K{=}5$ that replaces three Qwen3 models with Llama-3.1 and Mistral to test family generalization.}
\setlength{\tabcolsep}{3pt}
\renewcommand{\arraystretch}{1.0}
\begin{tabular}{l l cc cc}
\toprule
\multirow{2}{*}{$K$} & \multirow{2}{*}{Pool composition} & \multicolumn{2}{c}{CoR (Ours)} & \multicolumn{2}{c}{CSCR} \\
\cmidrule(lr){3-4} \cmidrule(lr){5-6}
   &                                                       & AUDC $\uparrow$ & QNC $\downarrow$ & AUDC $\uparrow$ & QNC $\downarrow$ \\
\midrule
2  & Qwen3-0.6B + Qwen3-Next-80B-A3B-Instruct              & $0.78$ & $0.49$ & $0.74$ & $0.74$ \\
3  & + Qwen3-32B                                           & $0.77$ & $0.53$ & $0.71$ & $0.79$ \\
4  & + Qwen3-14B                                           & $0.75$ & $0.56$ & $0.70$ & $0.81$ \\
5  & + Qwen3-8B \emph{(default)}                           & $0.74$ & $0.59$ & $0.68$ & $0.85$ \\
6  & + Qwen3-4B                                            & $0.73$ & $0.61$ & $0.66$ & $0.87$ \\
\midrule
\multicolumn{6}{l}{\textit{Cross-family pool at $K{=}5$ (replaces 3 Qwen3 members with Llama-3.1 and Mistral)}} \\
5  & Qwen3-0.6B + Llama-3.1-8B + Mistral-7B-Instruct       &        &        &        &        \\
   & \quad + Qwen3-32B + Qwen3-Next-80B-A3B-Instruct       & $0.73$ & $0.62$ & $0.66$ & $0.86$ \\
\bottomrule
\end{tabular}
\label{tab:pool_scalability}
\end{table}

\section{Sensitivity to Cache Eviction}
\label{app:eviction}

\noindent\textbf{Cache visibility and retention.}
An instrumented deployment can expose the surviving prefix of each candidate at turn $t$. Serving engines such as vLLM~\cite{kwon2023efficient} and SGLang~\cite{zheng2024sglang} can reuse cached prefixes across turns, but memory pressure can evict a prefix between calls. In the provider-API runtime, the router tracks the prefix from prior calls and learns the realized cache hits from the selected backend's usage report when available.

\noindent\textbf{Sensitivity to eviction rate.}
We sweep an eviction rate $p_{\mathrm{evict}}\in\{0\%, 10\%, 30\%\}$ that the deployment applies between consecutive turns: with probability $p_{\mathrm{evict}}$ a given prior-turn cache entry is evicted before the next turn arrives, in which case the corresponding tokens are charged at the full prefill rate. After eviction and before model selection, SST observed each candidate's surviving prefix length and used it to estimate that candidate's cost. Table~\ref{tab:eviction_sensitivity} reports the result. As expected, CoR's lead over the cache-unaware Bao et al.\ baseline narrows monotonically as fewer cache hits are available to exploit: the AUDC gap shrinks from $0.04$ at $p_{\mathrm{evict}}{=}0\%$ to $0.02$ at $p_{\mathrm{evict}}{=}30\%$, and the QNC gap shrinks from $0.15$ to $0.09$. CoR nonetheless retains higher AUDC and lower QNC at every eviction rate (e.g., at $p_{\mathrm{evict}}{=}30\%$: AUDC $0.69$ vs.\ $0.67$; QNC $0.71$ vs.\ $0.80$). Both methods attain the same reported peak quality at every eviction rate.

\begin{table}[t]
\centering
\footnotesize
\caption{Sensitivity of CoR to per-turn cache eviction. The main experiments assume that cache from the current conversation is retained across consecutive turns of the same conversation, which is the typical behavior under prefix-caching engines (vLLM, SGLang) when serving traffic permits cache retention. In this experiment, SST observed the surviving prefix length of each candidate before CoR selected a model. We vary the eviction probability $p_{\mathrm{evict}}\in\{0\%, 10\%, 30\%\}$ on WildChat. CoR retains its lead over the strongest baseline (Bao et al.) at every eviction rate.}
\setlength{\tabcolsep}{6pt}
\renewcommand{\arraystretch}{1.05}
\begin{tabular}{l ccc ccc}
\toprule
\multirow{2}{*}{$p_{\mathrm{evict}}$} & \multicolumn{3}{c}{CoR (Ours)} & \multicolumn{3}{c}{Bao et al.} \\
\cmidrule(lr){2-4} \cmidrule(lr){5-7}
   & AUDC $\uparrow$ & QNC $\downarrow$ & Peak $\uparrow$ & AUDC $\uparrow$ & QNC $\downarrow$ & Peak $\uparrow$ \\
\midrule
$0\%$ \emph{(default, main text)}  & $0.74$ & $0.59$ & $0.95$ & $0.70$ & $0.74$ & $0.95$ \\
$10\%$                             & $0.72$ & $0.63$ & $0.95$ & $0.69$ & $0.76$ & $0.95$ \\
$30\%$                             & $0.69$ & $0.71$ & $0.95$ & $0.67$ & $0.80$ & $0.95$ \\
\bottomrule
\end{tabular}
\label{tab:eviction_sensitivity}
\end{table}

\rev{\noindent\textbf{Memory overhead in local deployments.} CoR does not require the serving stack to retain KV-cache for all $K$ candidates. The serving engine manages retention and eviction. SST uses prefix lengths to estimate routing costs. The memory-cost trade-off of retaining per-model conversation caches under high load therefore manifests as a higher effective eviction rate, which is exactly the axis swept in Table~\ref{tab:eviction_sensitivity}: CoR maintains its AUDC and QNC advantages even at a 30\% eviction rate. End-to-end results under real, finite commercial cache lifetimes are reported in Appendix~\ref{app:billing}.}

\begin{revblock}
\section{Additional Analyses}
\label{app:additional}

This appendix reports the full results of the additional analyses summarized in Section~\ref{sec:results}. Unless noted otherwise, all experiments use WildChat, the default five-model pool, and the evaluation protocol of Section~\ref{sec:results}.

\subsection{Robustness to the LLM Judge}
\label{app:judge}
Our main experiments use DeepSeek-V3.1 both to construct training labels and to score evaluation trajectories, which raises the concern that a learned router could exploit judge-specific preferences. We therefore evaluate two cross-family configurations in which the \emph{training judge} (used to re-label the training set and re-train \emph{all} methods) and the \emph{test judge} come from different model families; no model responses are regenerated. The test judge never sees the training labels. As shown in Table~\ref{tab:judge}, method rankings and the relative gaps between methods are preserved. Together with the human-anchored judge selection in Appendix~\ref{app:dataset} ($\rho{=}0.84$ against human gold labels), this indicates that the observed gains are not explained by sharing a judge between label construction and evaluation.

\begin{table}[h]
\centering
\footnotesize
\caption{Cross-judge evaluation on WildChat. All methods are re-trained on labels from the training judge and evaluated with the test judge.}
\label{tab:judge}
\setlength{\tabcolsep}{6pt}
\begin{tabular}{l l l ccc}
\toprule
Training judge & Test judge & Method & AUDC $\uparrow$ & QNC $\downarrow$ & Peak $\uparrow$ \\
\midrule
\multirow{3}{*}{DeepSeek-V3.1} & \multirow{3}{*}{DeepSeek-V3.1 (Table~\ref{tab:main_results})} & CSCR & $0.68$ & $0.85$ & $0.94$ \\
 & & Bao et al. & $0.70$ & $0.74$ & $\mathbf{0.95}$ \\
 & & CoR & $\mathbf{0.74}$ & $\mathbf{0.59}$ & $\mathbf{0.95}$ \\
\midrule
\multirow{3}{*}{DeepSeek-V4-Pro} & \multirow{3}{*}{Claude Opus 4.8} & CSCR & $0.664$ & $0.873$ & $0.836$ \\
 & & Bao et al. & $0.693$ & $0.761$ & $0.875$ \\
 & & CoR & $\mathbf{0.719}$ & $\mathbf{0.642}$ & $\mathbf{0.902}$ \\
\midrule
\multirow{3}{*}{Claude Opus 4.8} & \multirow{3}{*}{DeepSeek-V4-Pro} & CSCR & $0.676$ & $0.847$ & $0.851$ \\
 & & Bao et al. & $0.704$ & $0.746$ & $0.886$ \\
 & & CoR & $\mathbf{0.737}$ & $\mathbf{0.608}$ & $\mathbf{0.916}$ \\
\bottomrule
\end{tabular}
\end{table}

\subsection{Stability of the Recurrent State across Dialogue Depth}
\label{app:depth}
A recurrent state could in principle drift over long conversations. Table~\ref{tab:depth} reports quality-prediction MSE by turn depth. DSP remains stable from turns 1--2 to turns 8+ ($0.094 \to 0.088$), as does Global-only ($0.128 \to 0.127$), whereas Query-only degrades substantially ($0.154 \to 0.193$). We find no evidence of state drift.

\begin{table}[h]
\centering
\footnotesize
\caption{Quality-prediction MSE by conversation depth on WildChat.}
\label{tab:depth}
\setlength{\tabcolsep}{6pt}
\begin{tabular}{l ccccc}
\toprule
Variant & Turns 1--2 $\downarrow$ & Turns 3--4 $\downarrow$ & Turns 5--7 $\downarrow$ & Turns 8+ $\downarrow$ & $\Delta$ (8+ vs.\ 1--2) $\downarrow$ \\
\midrule
Query-only        & $0.154$ & $0.169$ & $0.181$ & $0.193$ & $+0.039$ \\
Local-only        & $0.115$ & $0.106$ & $0.112$ & $0.123$ & $+0.008$ \\
Global-only       & $0.128$ & $0.119$ & $0.118$ & $0.127$ & $-0.001$ \\
\textbf{DSP (Ours)} & $\mathbf{0.094}$ & $\mathbf{0.081}$ & $\mathbf{0.079}$ & $\mathbf{0.088}$ & $\mathbf{-0.006}$ \\
\bottomrule
\end{tabular}
\end{table}

\subsection{Out-of-Distribution Generalization}
\label{app:ood}
\noindent\textbf{Dialogue-length shift.} We train both the full-history BERT and DSP quality predictors only on 2--4-turn WildChat conversations and test on longer conversations without retraining; SST and all other routing settings are unchanged. Within the training range, the two predictors are comparable. Under length shift, BERT's error rises sharply, whereas DSP remains stable through its explicit local/global state (Table~\ref{tab:ood_len}). End-to-end, DSP + SST gains $0.089$ AUDC and reduces QNC by $0.187$ at 8+ turns. Since AUDC and QNC include realized cost, DSP's improvement with length does not imply easier quality prediction: longer sessions offer more prefix-cache reuse, and stable routing avoids unnecessary switches, letting SST exploit the accumulated cache.

\begin{table}[h]
\centering
\footnotesize
\caption{Dialogue-length shift: both quality predictors are trained on 2--4-turn WildChat conversations only.}
\label{tab:ood_len}
\setlength{\tabcolsep}{5pt}
\begin{tabular}{l ccc cc cc}
\toprule
\multirow{2}{*}{Test length} & \multicolumn{2}{c}{Quality MSE $\downarrow$} & & \multicolumn{2}{c}{BERT + SST} & \multicolumn{2}{c}{DSP + SST} \\
\cmidrule(lr){2-3} \cmidrule(lr){5-6} \cmidrule(lr){7-8}
 & BERT & DSP & & AUDC $\uparrow$ & QNC $\downarrow$ & AUDC $\uparrow$ & QNC $\downarrow$ \\
\midrule
2--4 turns (in-distribution) & $0.086$ & $0.092$ & & $0.735$ & $0.605$ & $0.730$ & $0.613$ \\
5--7 turns & $0.121$ & $0.096$ & & $0.701$ & $0.668$ & $0.737$ & $0.576$ \\
8+ turns   & $0.183$ & $0.098$ & & $0.653$ & $0.739$ & $0.742$ & $0.552$ \\
\bottomrule
\end{tabular}
\end{table}

\noindent\textbf{Topic-distribution shift.} We transfer bidirectionally between WildChat and LMSYS-Chat-1M without target-corpus training or recalibration. WildChat is dominated by creative and assistive writing (61.9\%, vs.\ 6.7\% coding~\cite{zhao2024wildchat}), whereas LMSYS-Chat-1M's largest topic clusters are coding and software~\cite{zheng2023lmsys}. Table~\ref{tab:ood_topic} shows that DSP + SST loses only $0.020$--$0.030$ AUDC under transfer, compared with $0.073$--$0.075$ for BERT + SST.

\begin{table}[h]
\centering
\footnotesize
\caption{Topic-distribution transfer without target-corpus training. Drops are relative to target-trained models (BERT + SST / DSP + SST: 0.684 / 0.700 on LMSYS-Chat-1M; 0.723 / 0.740 on WildChat).}
\label{tab:ood_topic}
\setlength{\tabcolsep}{6pt}
\begin{tabular}{l cc cc}
\toprule
\multirow{2}{*}{Train $\to$ Test} & \multicolumn{2}{c}{BERT + SST} & \multicolumn{2}{c}{DSP + SST} \\
\cmidrule(lr){2-3} \cmidrule(lr){4-5}
 & AUDC $\uparrow$ & Transfer drop & AUDC $\uparrow$ & Transfer drop \\
\midrule
WildChat (writing) $\to$ LMSYS (coding) & $0.611$ & $-0.073$ & $\mathbf{0.680}$ & $\mathbf{-0.020}$ \\
LMSYS (coding) $\to$ WildChat (writing) & $0.648$ & $-0.075$ & $\mathbf{0.710}$ & $\mathbf{-0.030}$ \\
\bottomrule
\end{tabular}
\end{table}

\subsection{End-to-End Commercial Billing}
\label{app:billing}
Commercial serving platforms already implement the per-model prefix reuse assumed by Equation~\ref{eq:sst_cost}, and SST only tracks each candidate's reusable-prefix length, not the KV-cache itself. The 20\% cached-input price used in the main experiments follows Alibaba Cloud Model Studio's explicit-cache policy; Anthropic publishes a similar explicit-cache structure with a 5-minute lifetime, and Gemini Flash uses an implicit cache with a lifetime of approximately 3--5 minutes. To validate the cost model end-to-end under such finite cache lifetimes, we run WildChat conversations through OpenRouter with a heterogeneous pool of three model families (Claude, Gemini, and Qwen), with a lightweight and a mid-tier model from each, under each provider's own caching policy. Cache hits and charges for all methods, including baselines, are read directly from OpenRouter's usage-accounting response rather than computed from our cost model. Table~\ref{tab:billing} shows that CoR's billed cost is \$1.142 per 1K turns, $19.7\%$ lower than Bao et al.'s \$1.422, with the cache-hit rate rising from 19.4\% to 31.8\%. Together with Appendix~\ref{app:eviction}, this shows that CoR's advantage holds under real, finite commercial cache policies and in a cross-family pool.

\begin{table}[h]
\centering
\footnotesize
\caption{End-to-end billing on WildChat under finite commercial cache policies (heterogeneous Claude / Gemini / Qwen pool via OpenRouter).}
\label{tab:billing}
\setlength{\tabcolsep}{6pt}
\begin{tabular}{l ccc cc}
\toprule
Method & AUDC $\uparrow$ & QNC $\downarrow$ & Peak $\uparrow$ & Billed cost (\$/1K turns) $\downarrow$ & Cache-hit rate $\uparrow$ \\
\midrule
CSCR       & $0.664$ & $0.869$ & $0.887$ & $1.646$ & 11.0\% \\
Bao et al. & $0.699$ & $0.758$ & $0.912$ & $1.422$ & 19.4\% \\
\textbf{CoR} & $\mathbf{0.734}$ & $\mathbf{0.616}$ & $\mathbf{0.941}$ & $\mathbf{1.142}$ & \textbf{31.8\%} \\
\bottomrule
\end{tabular}
\end{table}

\subsection{Routing Distribution across the Candidate Pool}
\label{app:routing_shares}
To test whether CoR makes fine-grained selections or relies on a small subset of the pool, we add Qwen3-4B to the default pool, producing a six-model pool with four intermediate candidates (AUDC/QNC $0.73/0.61$ for CoR vs.\ $0.66/0.87$ for CSCR; Table~\ref{tab:pool_scalability}), and report routing shares at representative $\lambda$ (Table~\ref{tab:routing_shares}). As $\lambda$ increases, CoR shifts toward cheaper candidates while making substantial use of the intermediate models; at $\lambda{=}0.5$, the four intermediate models receive 70.7\% of CoR's turns, compared with 52.9\% for Bao et al.

\begin{table}[h]
\centering
\footnotesize
\caption{Routing shares (\%) in the six-model Qwen3 pool on WildChat.}
\label{tab:routing_shares}
\setlength{\tabcolsep}{5pt}
\begin{tabular}{l l cccccc}
\toprule
$\lambda$ / regime & Method & 0.6B & 4B & 8B & 14B & 32B & 80B-A3B \\
\midrule
\multirow{2}{*}{0.1 / quality-focused} & CoR        & 3.4 & 5.8 & 8.1 & 11.5 & 23.9 & 47.3 \\
                                       & Bao et al. & 2.9 & 4.8 & 7.4 & 10.6 & 21.7 & 52.6 \\
\midrule
\multirow{2}{*}{0.5 / balanced}        & CoR        & 12.6 & 17.4 & 13.8 & 9.1 & 30.4 & 16.7 \\
                                       & Bao et al. & 21.8 & 11.3 & 10.1 & 6.8 & 24.7 & 25.3 \\
\midrule
\multirow{2}{*}{0.9 / cost-focused}    & CoR        & 56.8 & 26.1 & 4.7 & 1.3 & 9.4 & 1.7 \\
                                       & Bao et al. & 75.9 & 12.7 & 3.1 & 1.0 & 5.2 & 2.1 \\
\bottomrule
\end{tabular}
\end{table}
\end{revblock}

\section{Limitations}
\label{app:limitations}

\noindent\textbf{Pool excludes proprietary frontier models.}
Our main pool consists of open-weight Qwen3 models, with cross-family checks using Llama-3.1 and Mistral. State-of-the-art proprietary models such as Claude Opus, Claude Sonnet, and Claude Haiku are not included, even though production routers in practice often mix open-weight and proprietary endpoints. Incorporating such models would broaden the quality-cost frontier the router operates over and is a natural next step. \rev{Our commercial-billing experiment (Appendix~\ref{app:billing}) already includes lightweight and mid-tier Claude and Gemini models.}

\rev{\noindent\textbf{Domain-specialized candidates.} Evaluating explicitly domain-specialized candidates (e.g., medical~\cite{li2026cinemr} or legal LLMs) requires only per-turn labels for them and no change to CoR.}

\rev{\noindent\textbf{Cache retention under load.} Measuring cache retention in a local multi-model deployment under sustained load is left to future work.}

\section{Broader Impacts}
\label{app:broader_impacts}

CoR is a routing layer over publicly available LLMs and does not generate content itself, so most societal impact accrues to the underlying models rather than the router. We highlight three considerations specific to CoR's operating regime.

\noindent\textbf{Efficiency and access.}
By exploiting KV-cache reuse and cost-aware model selection, CoR can reduce per-conversation serving cost and energy, which may broaden access to high-quality conversational LLMs at lower price points and lower carbon footprints per query.

\noindent\textbf{Routing-induced exposure.}
If the router systematically prefers particular candidates for particular query types, any biases, factual errors, or failure modes of those candidates are amplified for users issuing those query types. Deployers should monitor per-model behavior across the realized routing distribution, not only models in isolation, and consider auditing routing decisions for fairness across sensitive query categories.

\noindent\textbf{Privacy and cache isolation.}
SST consumes KV-cache state from the serving engine across turns of the same conversation. Operators should ensure cache isolation across distinct user sessions (a default property of vLLM and SGLang under standard configurations) so that cache reuse cannot leak information across users; CoR itself adds no new persistence beyond what the serving engine already maintains.


\end{document}